# Early warning of pedestrians and cyclists

Joerg Christian Wolf
*Innovation and Engineering Center California*
*Volkswagen Group of America, Inc.*
Belmont, CA, United States
ORCID: 0000-0001-5695-3490

*Abstract*— **State-of-the-art motor vehicles are able to break for pedestrians in an emergency. We investigate what it would take to issue an early warning to the driver so he/she has time to react. We have identified that predicting the intention of a pedestrian reliably by position is a particularly hard challenge. This paper describes an early pedestrian warning demonstration system.**

## I. Introduction

### A. Events Until Collision

If a pedestrian is about to cross the road it can be first recognized by his/her body language and most importantly his/her head pose [1]. The pedestrian observes the traffic in order to make a decision. If an approaching vehicle drives at constant speed in an urban scenario, the velocity seems to play no role in the decision making [2]. However, it does play a role if the driver is slowing down to waive the right of way. This happens very early before the crossing takes place. A complex interaction between driver and pedestrian takes place here. Schneemann and Gohl [3] has looked into detail how the "intention" of crossing can be recognized and what influences it. Now the pedestrian starts moving towards the curb and is actually stepping onto the road. This can be categorized as the "action" phase, see Figure 1. It is possible to recognize this by tracking the position of a pedestrian. It has been shown that object tracking alone is not sufficient if the pedestrian is still on the pavement [2]. However, if the pedestrian has not stepped onto the road yet, the reason for his trajectory is not always about crossing the road. One can easily see that uncertainty and errors increase, the earlier a prediction is made. Furthermore, "intention" recognition requires the examination of body language and is technically more complex than object tracking.

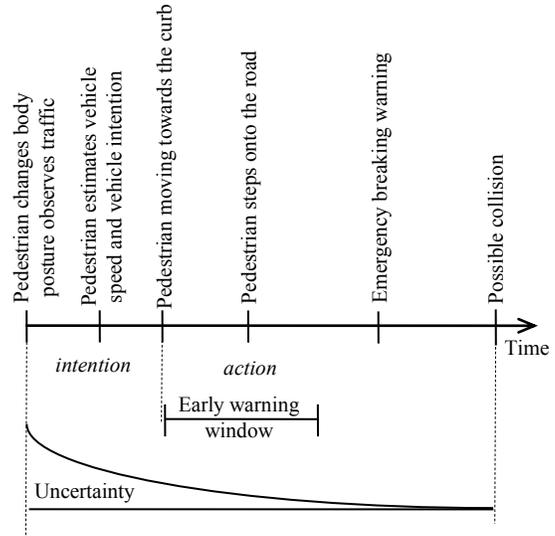

Fig. 1. Events until Collision, a timeline

In the next chapters we will describe an early warning system that relies on detecting the "action" of a pedestrian by position tracking.

### B. Combine Camera and navigation, great idea?

Initially we set out with the hypothesis that combining of camera and navigation technology in one system could create innovate new ways to guide the driver. A driver assistance camera that can detect pedestrians and other vehicles can make navigation guidance more context aware. We pursued ideas, such as telling the driver to follow a car that's turning and warn for pedestrians and bicycles. Such navigation prompts are a more natural user experience [4]. This paper describes a combined demonstration system. A voice prompt is issued to warn of pedestrians "Watch out for the pedestrian on the left". If a voice prompt was already playing or is about to be played the warning was combined with the maneuver: "turn right and watch out for bicycle on your right".

It turns out that navigation guidance information is issued well in advance to give the driver time to observe and plan. Current driver assistance systems, such as forward collision warning (FCW), pedestrian collision warning (PCW), blind spot detection (BSD) and lane keep assist (LKA), however, give the driver far less time to observe, plan and react. They operate on information in the immediate vicinity of the vehicle. Therefore the user experience concepts for driver assistance rely on lights, symbols and sounds rather than long voice prompts.

## II. SYSTEM SETUP

An Audi A7 was equipped with an additional automotive advanced driver assistance (ADAS) camera. The camera has the ability to detect other traffic participants, such as vehicles, bicycles and pedestrians. A PC received the position of observed objects from the camera via Controller Area Network (CAN) bus. Furthermore the wheel speed sensors, an accelerometer and gyro are sampled with 200 Hz and sent to a PC in the test car. The gyro, accelerometer, wheel sensors and a RTK GPS are integrated by Extended Kalman Filter in order to estimate the vehicle state. The state includes a global vehicle WGS position, yaw, pitch and roll as well as velocity. We have used partly external sensors in order to gain a more accurate position compared to the production vehicle.

In order to access the "most probable path" of the vehicle the PC requests a route using the HERE Maps [HERE ref] cloud routing engine. If the driver diverts from the route, a new route is calculated. The PC can also generate maneuver voice prompts, essentially forming a mini navigation system.

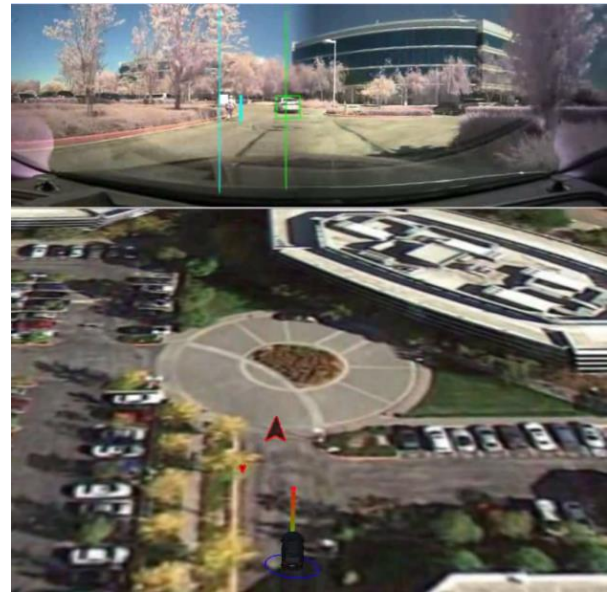

Fig. 3. Vehicle and positioning uncertainty at the bottom in a satellite image. At the top of the image is a stiched wide-angle camera image for recording with projected driver assistance camera detections.

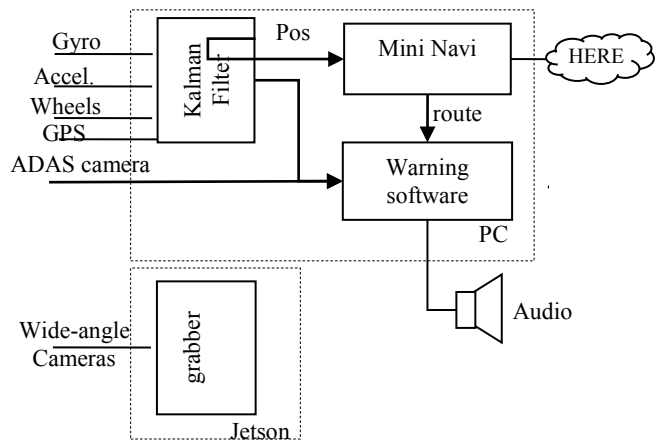

Fig. 2. System Overview

We have developed various visualizations of the observed data and vehicle including real-time satellite map overlay, an Android App with integrated HERE map, and a EB Assist ADTF (Automotive Data and Time-Triggered Framework) [adtf ref] rendered display.

## III. ACTION PREDICTION

### A. Algorithm

When a pedestrian appears in the camera his/her position is tracked at 36 samples a second to make "action" predictions of where the pedestrian is planning to go. The observed scene is transformed so that the vehicle faces the +x axis, using the vehicles initial heading. This allows for the pedestrian's longitudinal and lateral movement with respect to the vehicle to be analyzed on separate axes. A linear regression is applied to each axis with respect to time.

$$\hat{y}_{lat}(t) = \beta_{0lat}t + \beta_{1lat} \quad (1)$$

$$\hat{y}_{long}(t) = \beta_{0long}t + \beta_{1long} \quad (2)$$

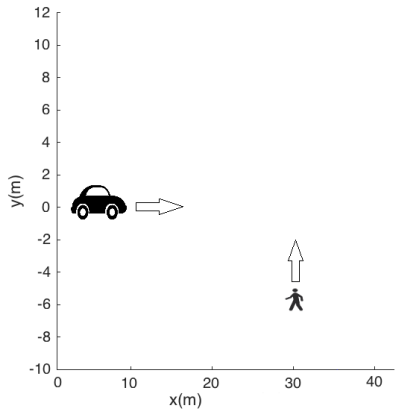

Fig. 4. vehicle moving longitudinal along x axis

A minimum number of samples is necessary to predict the pedestrian's motion. A monocular camera, can measure lateral positions much more accurately than longitudinal positions because it projects objects onto a assumed ground plane. Therefore lateral movement are calculated with a minimum of 12 samples (~333ms) while longitudinal movement require 30 samples (~833ms).

Having established a model for the pedestrian's motion it can be checked where the pedestrian's path is intersecting with the vehicles most probable path. Each pedestrian's interception point and time until interception is found by comparing the pedestrian's estimated trajectory with the vehicle's route. If the time until interception is less than 4 seconds and the interception point is at most 60 meters ahead of the vehicles's current position, then a warning is generated.

If the pedestrian is observed briefly before exiting the field of view (e.g. behind another car or the driver has passed by) and there are enough samples to estimate its trajectory, then the algorithm will continue to predict the pedestrian's location using its estimated trajectory for 4 seconds.

Pedestrian warnings provide directionality ("left", "right", "ahead") by estimating the pedestrian's position in the time it takes to hear the warning prompt. For example, if the warning prompt takes 1 second to read and a pedestrian crossing from left to right will be directly in front of the vehicle in 1 second then the algorithm will use "ahead".

We consider a warning as "early" if it can be issued at least 3 seconds before the emergency warning.

### B. Example Data

The following six figures illustrate pedestrian trajectories using data from the ADAS camera at different distances and vehicle speeds. In the first three figures, the vehicle is stationary at (0,0) facing the +x direction. In the second three, the vehicle is driving at 15mph in the +x direction starting at (0,0). In all six figures, the pedestrian is walking in the +y direction at x=20, 30 or 40 meters. A wide road and slow speed was chosen to give enough time to track the pedestrian through the complete field of view. This enabled us to investigate the tracking performance. We also conducted many test drives around urban and suburban towns whereby observed pedestrian would usually only have a track of 2-3 meters length.

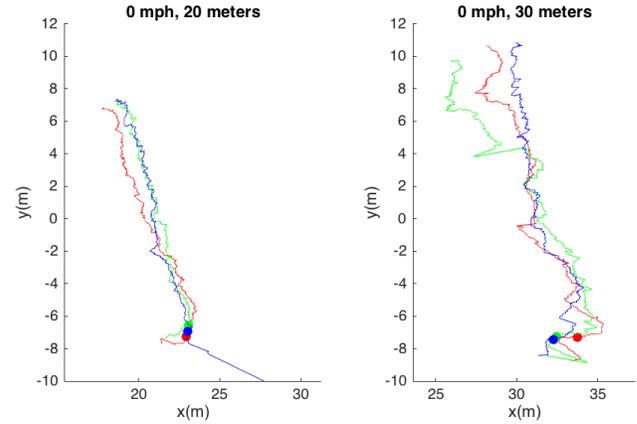

Fig. 5. Stationary vehicle at (0,0) pedestrian passes in +y direction

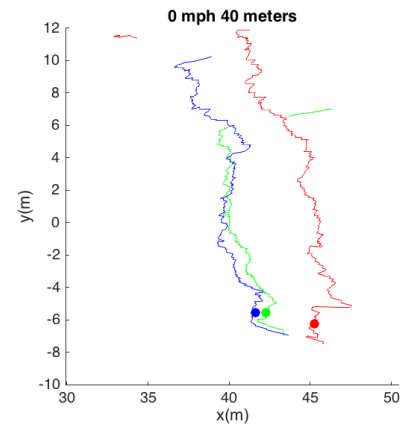

Fig. 6. Stationary vehicle at (0,0) pedestrian passes in +y direction, ground truth distance is 40 meters.

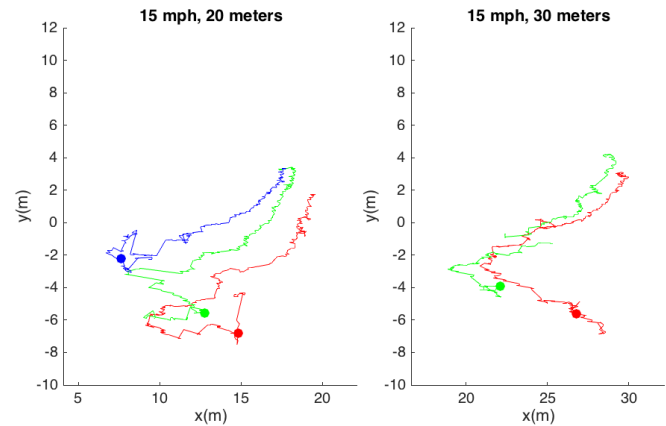

Fig. 7. Moving vehicle, motion compensated coordinates, pedestrian passes in +y direction

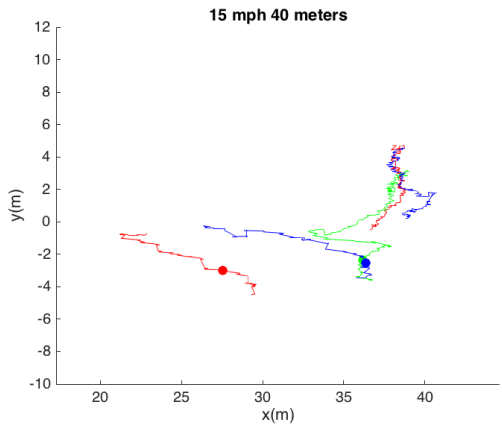

Fig. 8. Moving vehicle, motion compensated coordinates, pedestrian passes in +y direction, ground truth distance is 40 meters and is reducing as the vehicle gets closer.

The dot on each line marks the 30th sample of data. For almost every trajectory there is not enough position data up to the 30th sample to accurately estimate the direction in which the pedestrian is moving.

In each experiment with a stationary car the pedestrian trajectory is roughly linear. As expected, the start and end of each trajectory have the most noise while the center is the closest to the true position of the pedestrian (20, 30, and 40 meters, respectively). When the vehicle is moving, the ADAS camera struggles to locate the pedestrian. Figure 6 and figure 7 show that the ADAS camera both has trouble finding the initial position of the pedestrian and has a tendency to overcorrect for the motion of the car. As the car gets closer to the pedestrian, the position estimate gets more accurate until it converges on the true pedestrian position right before the pedestrian exists the camera field of view (FOV). This creates a sideways bell shaped curve. In figure 8, the ADAS camera's overcorrection is so severe that it creates a discontinuity in the trajectory.

It also occasionally loses track of pedestrians for a few milliseconds and then reclassifies them with a new ID. One pedestrian can therefore appear to be two or more because their trajectory is disjoint. In figure 8 the red disjoint trajectory is a single pedestrian. Each piece of the trajectory is relatively linear and mutually orthogonal. While there are some disjoint trajectories that can be merged with reasonable confidence, figure 8 shows that merging disjoint trajectories is not always possible. [5] has applied a particle filter to smooth the path of a tracked pedestrian. They reported that a change of one pixel can make the position jump several meters.

It is possible to consider a more complex model for the motion of a pedestrian and the car. A pedestrian never moves faster than a jogging pace and they are more likely to cross a street at an intersection. Since pedestrians move at walking speed then a max speed can be used to prevent erroneous trajectories calculated from data with a large amount of longitudinal error. [6] used Recursive Bayesian Filters which is a promising alternative.

Using this algorithm the system can distinguish between pedestrians walking along the pavement in longitudinal direction and pedestrians that are crossing and are from the timing in a collision course. The algorithm also works for bicycles. Most bicyclists are easy to monitor in low resolution ADAS cameras and can give correct trajectory estimations.

IV. DISCUSSION

A. Tracking Performance

To deliver a meaningful warning to the driver, the intention/action of pedestrians must be detected long before a potential collision. Pedestrian "action" is deduced primarily from their velocity. The camera however is very limited in its ability to track pedestrians and provide pedestrian position updates.

The most challenging problem is that current ADAS cameras typically have only a 40 degree FOV and are poor at estimating longitudinal distance of small objects. Its small FOV limits the number of pedestrian position samples before a potential collision and its ability to estimate longitudinal distance degrades with both true pedestrian distance and vehicle speed. Far more than 30 samples are typically needed to accurately estimate longitudinal movement. However, in most test drives, waiting for over 30 samples before warning about a pedestrian walking longitudinally took too long to be useful for an early warning.

Unlike longitudinal movement, lateral movement is much easier to detect. In most test drives, 12 samples were enough to accurately estimate lateral movement.

[7] argues that cameras still need to improve on stability, accuracy and range in order to reliably track the position of pedestrians. Currently it's only possible to get a limited field of view and far range or vice versa. We can confirm this with this experiment. It's not possible to reliably tell the direction in that a pedestrian is going using a camera, see figure 8.

Even more challenging is to recognize the head pose at range. For example [1] from another OEM argues that at the head needs to be at least 25 px width, which would require a 12 Mega-Pixel camera at a 40 degrees field of view to reach a 40m range.

B. User experience evaluation

Playing voice prompts had two mayor drawbacks: Firstly, if another prompt was already playing it delayed the warning. Secondly, if route guidance is not running it did not provide a coherent user experience with other driver assistance systems for pedestrians, which only make a warning sound. We now believe the user experience of the warning should be consistent with the HMI user experience of a pedestrian collision warning but should sound not so intense or urgent. We have considered playing the sound from the direction the threat is coming from.

## V. Opportunities For Future Work

The pedestrian's intent to cross can be partly determined on their location on the sidewalk. We observed from videos that only when the pedestrian is actually stepping from the pavement onto the road we can be sure he/she is trying to cross. This was not implemented in our test system, however it is technically feasible to use camera recognized lanes or HD map border geometry to determine when a pedestrian steps into a road.

Additionally, the driver's intent can be derived from the car's position data and if the driver has the foot on the breaks/accelerator. If the car slows down or is stationary it's possible that warnings aren't necessary. This would be another low-hanging fruit.

As seen in figure 1, we assume that uncertainty decreases as the pedestrian takes action and steps onto the road. We believe the best approach is a combination of intention recognition at first, i.e. head direction, etc. and then waiting for the action. This can be fused into a Bayesian statistical model, such as a hidden markov model or related. See for example the work from Kooji, et .al. [8].

We see potential that an "*intention*" recognition system based on a camera that is fused with an "*action*" recognition system that tracks movement with LIDAR/Radar can reduce uncertainty and make early warnings possible.